\documentclass[11pt,a4paper,table]{article}
\usepackage[hyperref]{acl2017}

\usepackage{times}
\usepackage{latexsym}

\usepackage{lipsum,adjustbox}
\usepackage{multicol, blindtext}

\usepackage{url}

\usepackage{pgfplots}
\usepackage{subcaption}

\usepackage[subtle]{savetrees} 
\usepackage{enumitem,kantlipsum}

\usepackage{verbatim} %provides the 'comment' environment

\aclfinalcopy % Uncomment this line for the final submission
\definecolor{sRed}{HTML}{C11B17}
\definecolor{sMagenta}{HTML}{d33682}
\definecolor{sViolet}{HTML}{6c71c4}
\definecolor{sGreen}{HTML}{859900}
\definecolor{sYellow}{HTML}{b58900}
\definecolor{sOrang}{HTML}{cb4b16}
\definecolor{sBlue}{HTML}{268bd2}
\definecolor{sCyan}{HTML}{2aa198}

\usepackage[utf8]{inputenc}
\usepackage{xargs}                      % Use more than one optional parameter in a new commands

\usepackage[colorinlistoftodos,prependcaption,textsize=tiny]{todonotes}
\newcommandx{\unsure}[2][1=]{\todo[linecolor=red,backgroundcolor=red!25,bordercolor=red,#1]{#2}}
\newcommandx{\change}[2][1=]{\todo[linecolor=blue,backgroundcolor=blue!25,bordercolor=blue,#1]{#2}}
\newcommandx{\info}[2][1=]{\todo[linecolor=OliveGreen,backgroundcolor=OliveGreen!25,bordercolor=OliveGreen,#1]{#2}}
\newcommandx{\improvement}[2][1=]{\todo[linecolor=Plum,backgroundcolor=Plum!25,bordercolor=Plum,#1]{#2}}
\newcommandx{\thiswillnotshow}[2][1=]{\todo[disable,#1]{#2}}

\title{Experiential, Distributional and Dependency-based Word Embeddings have Complementary Roles in Decoding Brain Activity}

\author{
Samira Abnar 
\(\quad\)
Rasyan Ahmed
\(\quad\)
Max Mijnheer
\(\quad\)
Willem Zuidema
\(\quad\) \\
University of Amsterdam \\
{\tt \{samiraabnar,rasyan21,max.mijnheer\}@gmail.com, zuidema@uva.nl}\\ }

\date{}

\begin{document}

\maketitle

\begin{abstract}
We evaluate 8 different word embedding models on their usefulness for predicting the neural activation patterns associated with concrete nouns. The models we consider include an experiential model, based on crowd-sourced association data, several popular neural and distributional models, and a model that reflects the syntactic context of words (based on dependency parses). Our goal is to assess the cognitive plausibility of these various embedding models, and understand how we can further improve our methods for interpreting brain imaging data.

We show that neural word embedding models exhibit superior performance on the tasks we consider, beating experiential word representation model.
The syntactically informed model gives the overall best performance when predicting brain activation patterns from word embeddings; whereas the GloVe distributional method gives the overall best performance when predicting in the reverse direction (words vectors from brain images). Interestingly, however, the error patterns of these different models are markedly different. This may support the idea that the brain uses different systems for processing different kinds of words. Moreover, we suggest that taking the relative strengths of different embedding models into account will lead to better models of the brain activity associated with words.
\end{abstract}

\section{Introduction}
How are word meanings represented in the human brain? Is there a single amodal semantic system or are there multiple responsible for representing meanings of different classes of words? 
Recently, a series of studies have emerged showing that a combination of methods from machine learning, computational linguistics and cognitive neuroscience are useful for addressing such questions.

\cite{mitchell2008predicting} pioneered the use of corpus-derived word representations to predict patterns of neural activation's when subjects are exposed to a stimulus word. Using their framework, a series of papers have evaluated various techniques of computing word representation models based on different assumptions, as we review in section~\ref{relatedwork}. 

Since these early successes, a range of new word embedding methods have been proposed and successfully used in a variety of NLP tasks, including methods based on deep learning with neural networks. \cite{baroni2014don} and \cite{pereira2016comparative} present systematic studies, showing that also behavioural data from psycholinguistics can be modelled effectively using neural word embedding models such as GloVe\cite{pennington2014glove} and word2vec\cite{word2vec1}. At the same time, studies in the area of vision have shown that deep learning models fit very well to the neocortical data \cite{cadieu2014deep,khaligh2014deep} and they can help to better understand the sensory cortical system \cite{yamins2016using}. To investigate how well the new word embedding models, and in particular the deep learning models, fare in helping to understand neural activation patterns in the domain of language, we now present a systematic evaluation of 8 word embedding models, listed in section~\ref{experimental-setup}, against the neuroimaging data from \cite{mitchell2008predicting}, following the experiments and primary results in \cite{max:thesis,rasyan:thesis}.

\newcommand{\maingoal}{to evaluate the usefulness of these different word embedding models for understanding different properties and features reflected in human brain neural activation patterns. }

\newcommand{\RQOne}{\textit{How well does each word embedding model allow us to predict neural activation patterns in human brain?}}
\newcommand{\RQTwo}{\textit{Which are the most predictable voxels in the brain for each word embedding model? }}
\newcommand{\RQThree}{\textit{What is the best word embedding model for predicting brain activation for different (classes of) nouns? }}

To address this goal, we take word embedding models designed based on different assumptions of how meanings of words can be represented and evaluate their performance on either the task of predicting brain data from word embeddings or the reverse, predicting word embeddings from brain data.
The basic assumption here is that the better the performance of a model is the more probable it is that the way the word embedding model is built reflects what happens in the human brain to understand a meaning of a word.
In our experiments, we compare modern neural word embedding models with traditional approaches that are based on manually assigned linguistic word attributes, and neuro-inspired techniques based on sensory-motor features.
Besides a large-scale evaluation of various word embedding models, we conduct a detailed error analysis to understand the differences between them.

The first research question we investigate is:
\RQOne\
To answer this we measure how well different word embedding models can predict the brain imaging data. Taking this one step further, we also train our models in the reverse direction: to directly predict word embeddings from brain data.

The second research question that we investigate is:
\RQThree\
Maybe human brain uses different processes to understand meanings of different kind of words \cite{riddoch1988semantic,caramazza1990multiple,warrington1984category,caramazza1998domain}. We do a qualitative analysis of our results to see whether different word embedding models are good in predicting the brain activation for different categories of nouns.
The third question we address is
\RQTwo\
By answering this question we want to test the hypothesis that different areas of the brain are responsible for processing different aspect of the meaning of nouns.
If different models have different performance either for different noun pairs or for different brain areas, the next step would be to find a way to integrate different models to build a model that better fits the brain data.

\section{Related Work}
\label{relatedwork}

The tradition of developing computational models to predict neural activation patterns given a representation of a stimulus such as a word was started by \cite{mitchell2008predicting}, who presented a model that quite successfully (with performance well above chance) predicted neural activation patterns associated with nouns, using a hand-designed set of 25 verbs (reflecting sensory-motor features) and computing representations for the nouns based on their co-occurrences with these verbs in a trillion-token corpus.
Following this work, \cite{jelodar2010wordnet} proposed using WordNet \cite{miller1995wordnet} instead of corpus statistics to compute the values for the 25 features introduced in \cite{mitchell2008predicting}, allowing them to deal with some of the ambiguity related issues.  They find that a linear combination of their WordNet-based 25 features and the co-occurrence based 25 features of \cite{mitchell2008predicting} improves the fMRI neural activity prediction accuracy.
Devereux et al \cite{devereux2010using} applied the framework to evaluate four different feature extraction methods, each based on a different source of information available in corpora. They show that general computational word representation models can be as good as sensory-motor based word representations. 
Later Murphy et al have done an extensive study comparing the performance of a different kind of corpus-based models on this task. In their experiments, a model that exploits dependency information outperforms the others \cite{murphy2012selecting}, in line with the results that we report below.
\cite{binder2016toward} argue that it makes more sense to use experiment based word representations to model the mental lexicon.
In \cite{fernandino2015predicting} they use sensory-motor experience based attributes as elements of the word vectors to predict neural activation pattern for lexical concepts. The main difference of this approach with \cite{mitchell2008predicting} is that rather than statistics from corpora they use actual human ratings to compute the feature values.

More recently, the success of neural network based approaches for learning word representations has raised the question whether these models might be able to partly simulate how our brain is processing language. 
Hence, it is now the time to revisit the challenge Tom Mitchell introduced and evaluate these new models with human brain neural activation patterns.
In \cite{anderson2017visually} the performance of word2vec as the word representation model for predicting brain activation patterns is already evaluated. The goal of their experiment was to compare a text-based word representation with image-based models; our goal, instead, is to compare different neural word embedding models that are all text-based. 
Furthermore, \cite{xu2016brainbench} they compare the performance of various word embedding models, including neural based models and non-distributional models for both behavioural tasks and brain image datasets.

Taking the differences between all these different models for word representation into account, one can argue that they are not replaceable with each other.
In \cite{dove2009beyond} it is argued that both perceptual and non-perceptual features are important in decoding semantics. 
Moreover \cite{andrews2009integrating} has suggested combining experiential and distributional models to learn word representations. 
In our experiments, we want to investigate whether the information encoded in different kind of word representations are mutually exclusive and hence, integrating them would result in a more powerful model.

There have also been some efforts to extend these models to analyze and understand brain activation patterns at sentence level \cite{wehbe2014simultaneously} or at least in the context of a sentence rather than an isolated word \cite{anderson2016predicting}. Moreover, some other related work abstracts away from the brain activation patterns and instead analyzes the correlation between the pairwise similarity of word representations in the brain and the computational model under evaluation \cite{anderson2016representational}.

In this paper, we stay with the original setup, using word representation models for predicting fMRI neural activation patterns, but go beyond existing work by presenting a systematic analysis and comparison of the performance of different kind of word representation models.

\section{Experimental Setup}
\label{experimental-setup}

The main task in our experiments is to use a regression model to map word representations to brain activation patterns or vice versa.
As the regression model, we employ a single layer neural network with $tanh$ activation. 
To avoid over-fitting we use drop-connect \cite{wan2013regularization} with a keeping rate of 0.7 beside L2 regularization with $\lambda = 0.001$. 
In all the experiments we train the models for each subject separately.
The training and evaluation are done with the leave-2-out method as suggested in \cite{mitchell2008predicting}. Where we train the model on all except 2 pairs and then evaluate the performance of the model on the left-out pairs. 
We do this for all possible combinations of pairs. 

\paragraph{Neuroimaging Data}
Our experiments are conducted on the data from \citet{mitchell2008predicting} which is publicly available\footnote{\url{http://www.cs.cmu.edu/afs/cs/project/theo-73/www/science2008/data.html}}.
This is a collection of fMRI data that is gathered from 9 participants while exposed to distinctive stimuli. The stimuli consisted of 60 nouns and corresponding line drawings. Each stimulus was displayed six times for 3 seconds in random order, adding to a total of 360 fMRI images per participant. 
\paragraph{Word Embedding Models}
In order to get insights about how human mental lexicon is built, we use a wide variety of recently proposed word representation models.
%These models are built based on different hypothesis.
The word embedding models that we are exploring in our experiments are in two (non-exclusive) categories: experiential or distributional.
In the experiential model, the meanings of the words are coded to reflect how the corresponding concept is experienced by humans through their senses. 
In the distributional models, the meaning of words is represented based on their co-occurrence with other words. These models can be either count-based or predictive \cite{baroni2014don}.
The word representation models we will use are:
\begin{itemize}[leftmargin=*]
    \item \textbf{Experiential word representations}:
    Experiential word representations are suggested based on the fact that humans remember the meaning of things as they experience them. In \cite{binder2016toward} a set of 65 features are defined and crowdsourcing is used to rate the relatedness of each feature for each word. Thus, instead of computing the value of features using statistical data from textual corpora they use actual human ratings.  We use the dataset introduce in \cite{binder2016toward}. Since it contains only about $50\%$ of the nouns in Tom Mitchell et al dataset, some of the experiments we report are with this limited noun set.
    \item \textbf{Distributional word embedding models}:
    \begin{itemize}[leftmargin=*]
        \item \textbf{Word2Vec}: Word2vec basically is a shallow, two layer, neural network that reconstructs the context of a given word. In our experiments, we use the skip gram word2vec model trained on Wikipedia \cite{word2vec1}.
        \item \textbf{Fasttext}: Fasttext is a modification of word2vec that takes morphological information into account \cite{bojanowski2016enriching}. 
        %For this purpose, in this model, words are represented as a bag of character n-gram.
        \item \textbf{Dependency-based word2vec}: The dependency-based word2vec introduced in \cite{levy2014dependency} is a word2vec model in which the context of the words is computed based on the dependency relations.
        \item \textbf{GloVe}: 
        GloVe is a count-based method. It does a dimensionality reduction on the co-occurrence matrix\cite{pennington2014glove}.
        \item \textbf{LexVec}: 
        LexVec is also a count based method. It is a matrix factorization method that combines ideas from different models. It minimizes the reconstruction loss function that weights frequent co-occurrences heavily while taking into account negative co-occurrence \cite{salle2016matrix,salle2016enhancing}.
    \end{itemize}
    \item \textbf{25 verb features}: 
    Similar to experiential word representations, this model is based on the idea that the neural representation of nouns is grounded in sensory-motor features. They have manually picked 25 verbs and suggested to use the co-occurrence counts of nouns with these 25 verbs to form the word representations \cite{mitchell2008predicting}. 
    \item \textbf{non-distributional word vector representation}: 
    \cite{faruqui:2015:non-dist} have constructed a non-distributional word representation model employing linguistic resources such as WordNet\cite{miller1995wordnet}, FrameNet\cite{baker1998berkeley} etc. In this model, words are presented as binary vectors where each element of the vector indicates whether the represented word has or does not have a specific feature. As a result, the vectors are highly sparse. The advantage of this model to distributional word representations is the interpretability of its dimensions. 
\end{itemize}

\begin{figure}[t]
    \centering
    \begin{tikzpicture}
\centering
\begin{axis}[
	symbolic x coords={30 words,60 words},
	xtick = data,
	ylabel= Accuracy,
	xlabel = Averages over 9 subjects,
	enlarge x limits=0.6,
	legend style={legend image post style={xscale=0.5}, at={(0.5,1.6),font=\fontsize{7}{8}\selectfont},
   legend columns=3, anchor= north,
   },
   area legend,
	ybar,
	ymin=0.5,
    ymax=1.0, 
    point meta={y*100},
	bar width=6pt,
	width  = 0.48 * \textwidth,
    height = 4.0cm,
    major x tick style = transparent,
    ymajorgrids,
    minor tick num=1,
    ytick = {0.1,0.3,0.5,0.7,0.9},
    ticklabel style = {font = \fontsize{9}{10}\selectfont},
    label style = {font = \fontsize{10}{11}\selectfont},
    major x tick style = transparent,
    y label style={at={(axis description cs:0.06,.5)}},
]
%experiential
\addplot [fill=sMagenta, draw=sMagenta]
	coordinates { (30 words, 0.75) (60 words,0.0)};
	
%F25
\addplot  [fill=sCyan, draw=sCyan]
	coordinates {(30 words, 0.72) (60 words,0.78)};
%Glove 
\addplot [fill = sBlue, draw = sBlue]
	coordinates {(30 words, 0.74) (60 words,0.79)};
%Fasttext
\addplot [fill = sYellow, draw = sYellow]
	coordinates {(30 words, 0.74) (60 words,0.77)};
%word2vec
\addplot  [fill = sRed, draw = sRed]
	coordinates { (30 words, 0.77) (60 words,0.73)};
		 
%dep based
\addplot [fill = sOrang, draw = sOrang]
	coordinates {(30 words, 0.81) (60 words,0.80)};

%lexvec
\addplot  [fill = sGreen, draw = sGreen]
	coordinates {(30 words, 0.77) (60 words, 0.79)};
		 
%non dist		 
\addplot [fill = sViolet, draw = sViolet]
	coordinates { (30 words, 0.70) (60 words,0.71)};

\legend{experiential, 25 features,Glove,FastText, Word2Vec, Dependency based, lexvec, Non Distributional}
\end{axis}
\end{tikzpicture}
    \vspace{-10pt}
    \caption{Results for the word to brain activation prediction task. (Chance is .5)}
    \label{fig:Exp1_allinall}
\end{figure}
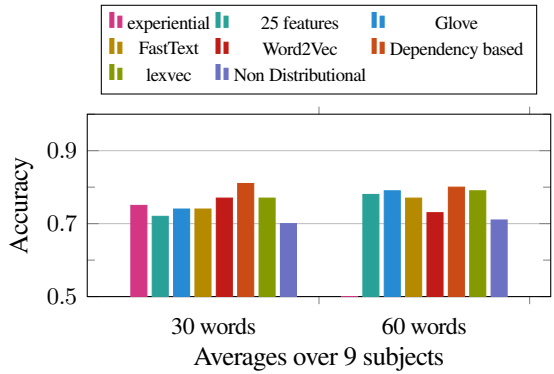

\begin{figure*}[htbp]
    \centering
    \begin{tikzpicture}
\begin{axis}[
	symbolic x coords={1,2,3,
		4, 5,6,7,8,9, avg},
	xtick = data,
	ylabel= Accuracy,
	xlabel = Subjects,
	enlargelimits=0.05,
	legend style={legend image post style={xscale=0.5}, at={(0.5,1.30),font=\fontsize{7}{8}\selectfont},
   legend columns=-1, anchor= north,
   },
   area legend,
	ybar,
	ymin=0.5,
    ymax=1.0, 
    point meta={y*100},
	bar width=2pt,
	width  = \textwidth,
    height = 4cm,
    major x tick style = transparent,
    ymajorgrids,
    minor tick num=1,
    ytick = {0.1,0.3,0.5,0.7,0.9},
    ticklabel style = {font = \fontsize{8}{9}\selectfont},
    label style = {font = \fontsize{10}{11}\selectfont},
    major x tick style = transparent,
]
%25F
\addplot [fill=sCyan, draw=sCyan]
	coordinates {(1,0.78) (2,0.72)
		 (3,0.79) (4,0.81) (5,0.74) (6,0.59)
    	 (7,0.71) (8,0.55) (9,0.83) (avg, 0.72)};
%experiential
\addplot [fill=sMagenta, draw=sMagenta]
	coordinates {(1,0.89) (2,0.64)
		 (3,0.77) (4,0.90) (5,0.74) (6,0.69)
		 (7,0.75) (8,0.52) (9,0.89) (avg, 0.75)};
%glove
\addplot [fill = sBlue, draw = sBlue]
	coordinates {(1,0.86) (2,0.54)
		 (3,0.82) (4,0.76) (5,0.71) (6,0.73)
		 (7,0.76) (8,0.68) (9,0.76) (avg, 0.74)};
%fasttext
\addplot  [fill = sYellow, draw = sYellow]
	coordinates {(1,0.89) (2,0.70)
		 (3,0.78) (4,0.87) (5,0.65) (6,0.68)
		 (7,0.80) (8,0.55) (9,0.78) (avg, 0.74)};
%word2vec
\addplot  [fill = sRed, draw = sRed]
	coordinates {(1,0.87) (2,0.78)
		 (3,0.81) (4,0.75) (5,0.83) (6,0.71)
		 (7,0.74) (8,0.64) (9,0.82) (avg, 0.77)};

%dep based		 
\addplot  [fill = sOrang, draw = sOrang]
	coordinates {(1,0.94) (2,0.74)
		 (3,0.86) (4,0.92) (5,0.80) (6,0.71)
		 (7,0.87) (8,0.61) (9,0.83) (avg, 0.81)};
		 
%lexvec
\addplot  [fill = sGreen, draw = sGreen]
	coordinates {(1,0.88) (2,0.71)
		 (3,0.82) (4,0.88) (5,0.71) (6,0.72)
		 (7,0.72) (8,0.66) (9,0.83) (avg, 0.77)};

%non dist		 
\addplot  [fill = sViolet, draw = sViolet]
	coordinates {(1,0.81) (2,0.58)
		 (3,0.74) (4,0.76) (5,0.68) (6,0.69)
		 (7,0.64) (8,0.54) (9,0.84) (avg, 0.70)};

\legend{25 features, Experiental,Glove,FastText, Word2Vec, Dependency based,lexvec, Non Distributional}
\end{axis}
\end{tikzpicture}
    \caption{Results of different word representation models for the word to brain activation prediction task for the limited set of word, split per subject.}
    \label{fig:Exp1_allinall_limited}
\end{figure*}
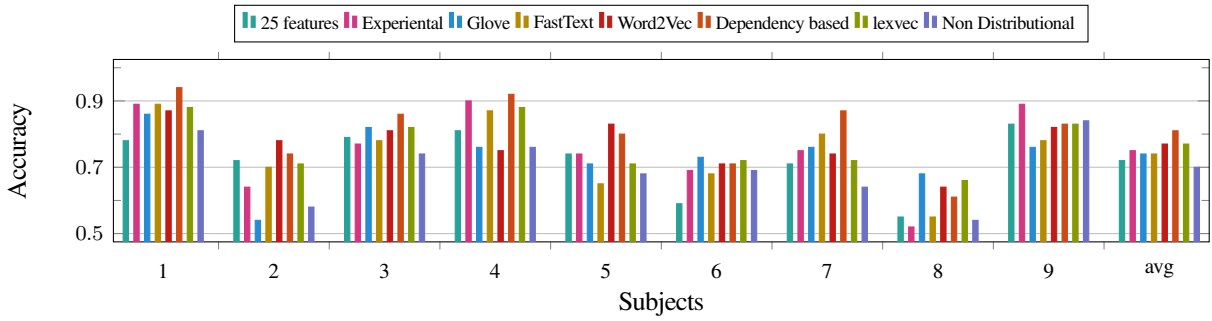

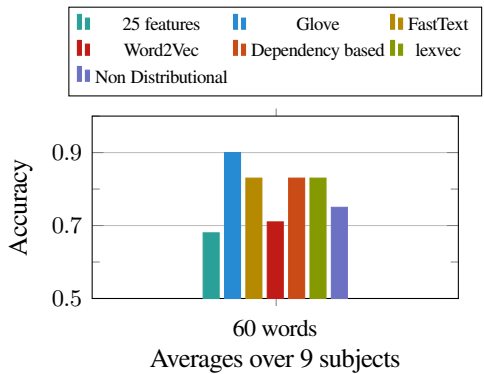
\begin{figure}[ht]
    \centering
    \begin{tikzpicture}
\centering
\begin{axis}[
	symbolic x coords={60 words},
	xtick = data,
	ylabel= Accuracy,
	xlabel = Averages over 9 subjects,
	enlarge x limits=.8,
	legend style={legend image post style={xscale=0.5}, at={(0.5,1.6),font=\fontsize{7}{8}\selectfont},
   legend columns=3, anchor= north,
   },
   area legend,
	ybar,
	ymin=0.5,
    ymax=1.0, 
    point meta={y*100},
	bar width=6pt,
	width  = 0.4 * \textwidth,
    height = 4.0cm,
    major x tick style = transparent,
    ymajorgrids,
    minor tick num=1,
    ytick = {0.1,0.3,0.5,0.7,0.9},
    ticklabel style = {font = \fontsize{9}{10}\selectfont},
    label style = {font = \fontsize{10}{11}\selectfont},
    major x tick style = transparent,
    y label style={at={(axis description cs:0.06,.5)}},
]
	
%F25
\addplot  [fill=sCyan, draw=sCyan]
	coordinates { (60 words,0.68)};
%Glove 
\addplot [fill = sBlue, draw = sBlue]
	coordinates { (60 words,0.90)};
%Fasttext
\addplot [fill = sYellow, draw = sYellow]
	coordinates { (60 words,0.83)};
%word2vec
\addplot  [fill = sRed, draw = sRed]
	coordinates {  (60 words,0.71)};
		 
%dep based
\addplot [fill = sOrang, draw = sOrang]
	coordinates { (60 words,0.83)};

%lexvec
\addplot  [fill = sGreen, draw = sGreen]
	coordinates { (60 words, 0.83)};
		 
%non dist		 
\addplot [fill = sViolet, draw = sViolet]
	coordinates {  (60 words,0.75)};

\legend{25 features,Glove,FastText, Word2Vec, Dependency based, lexvec, Non Distributional}

\end{axis}
\end{tikzpicture}
    \vspace{-10pt}
    \caption{Results of different word representation models for the brain activation to word representation prediction task.}
    \label{fig:Exp2_allinall}
\end{figure}

\section{\RQOne}
To address the first research question, we train a separate regression model for each word representation model to compute the average brain activation corresponding to each word for a particular subject. Figure \ref{fig:Exp1_allinall} illustrates the results of evaluating these models on the brain activation prediction task, using the leave-2-out methodology we discussed in section~\ref{experimental-setup}.
For the sake of including the experiential word representations from \cite{binder2016toward} in our evaluations, we also conducted a set of experiments with only the nouns that were included in the experiential word representation collection.
The good news is that all the models we are evaluating perform significantly above chance. The fact that the ranking of the models differs per subject makes it difficult to make general conclusions about the best model. Overall, dependency-based word2vec, GloVe and 25 features model are the top-ranked models for at least one of the subjects. 

Among neural word embedding models, dependency-based word2vec is achieving the best accuracy. This is in line with the results from \cite{murphy2012selecting}, where they showed that the corpus-based model considering the dependency relationships has the highest performance among corpus-based models. These authors report an accuracy of 83.1 (with 1000 dimensional word vectors). Somewhat higher still than the best dependency based word2vec, and the highest performance reported in the literature until now for a corpus-based model. 
The fact that fasttext and dependency based word2vec are performing better than word2vec might reflect the importance of morphological and dependency information.
%Comparing, word2vec, fasttext, and dependency based word2vec, fasttext and dependency based word2vec are performing better. This can prove the importance of morphological and dependency information.
Comparing predictive models with count-based models, although count-based methods like GloVe and LexVec are beating simple word2vec, looking at the performances of fasttext and dependency based word2vec, we can conclude that the context prediction models can potentially perform better.
%As the results shown in Figure \ref{fig:Exp1_allinall} indicate, the dependency based word2vec achieves the highest accuracy also on this limited set of nouns.
Moreover, comparing the performance of the Experiential Model with 25 feature model, we see that the Experiential Model is doing slightly better on average while their ranking is different per subject. Either the higher number of features or the way feature values are computed could have led to the slight improvement in accuracy for the experiential model. 

In both sets of experiments in Figure \ref{fig:Exp1_allinall} the non-distributional word representation model has the lowest performance. The very high dimensionality of the brain imaging data versus the sparseness of non-distributional word vectors make training the regression model with these vectors much harder and this might be the primary reason for its low performance.

Next, instead of predicting brain activation patterns, we train the regression model to predict the word representation given a brain activation. 
Thus, we want to predict the stimulus word from the neural activation pattern in the brain. %Or we are decoding brain activation patterns.
Evaluation is still based on the leave-2-out setup (so we still evaluate with 2 brain images and 2 word embeddings at each instance, making quantitative results comparable across experiments).

The results are shown in Figure \ref{fig:Exp2_allinall}.
%Comparing the results from Figures \ref{fig:Exp1_allinall} and \ref{fig:Exp2_allinall} gives us an impression of how expressive each word representation model is (including neural activation patterns).
We expected the performance of the models on the reversed task, predicting word features from brain activation, to be somewhat similar to their performance on the main task, predicting brain activation patterns from word vectors. However, the results are surprising. 
%For all models except the 25 features model the accuracy on the reversed task is a bit higher. This is reasonable if consider the logical assumption that brain representations are more general than the computational model.
For the 25 features model, the accuracy on the reversed task is much lower. This may be because of the way the feature vector for nouns is distributed in the space in this model. Or it could be that neural activation patterns do not encode all the necessary information to approximate these feature values. This could indicate that while the 25 features model is pretty useful in interpreting brain activation patterns it is not a plausible model to simulate how nouns are represented in the human brain. On the other hand, it seems that it is very easy to construct GloVe word vectors from brain activation patterns; this model achieves an accuracy of 90 percent. In \cite{sudre2012tracking} accuracy of 91.19 percent is reported on the similar task on MEG data.
GloVe is based on the distributional semantics hypothesis, and it is achieved by learning to predict the global co-occurrence statistics of words in a corpus. Hence, obtaining a high accuracy in the word prediction task using GloVe, supports the fact that the context of the words have a major role in the way we learn the meanings of the words. The important thing to notice is that of course the more information we encode in the word representation the more powerful it becomes in predicting neural activation patterns as far as that information are relevant to some extent. However, this alone doesn't imply that the exact same information is encoded in the neural activation patterns. As we can see in our results, compared to GloVe, it's not that easy to reconstruct the Fasttext and dependency based word vectors from the brain activation patterns. What we can conclude, for now, is that while morphological and dependency information is helpful in learning word representations that are to some extent more similar to the neural representation of nouns in our brain. This information is not explicitly encoded in the brain activation patterns.

In the end, only comparing the accuracy of these models does not reveal much about the differences between them and does not mean that the model with the highest accuracy can replace all the others. 

\begin{figure}
        \includegraphics[width=.5\textwidth,trim={0 0 0 1.3cm}]{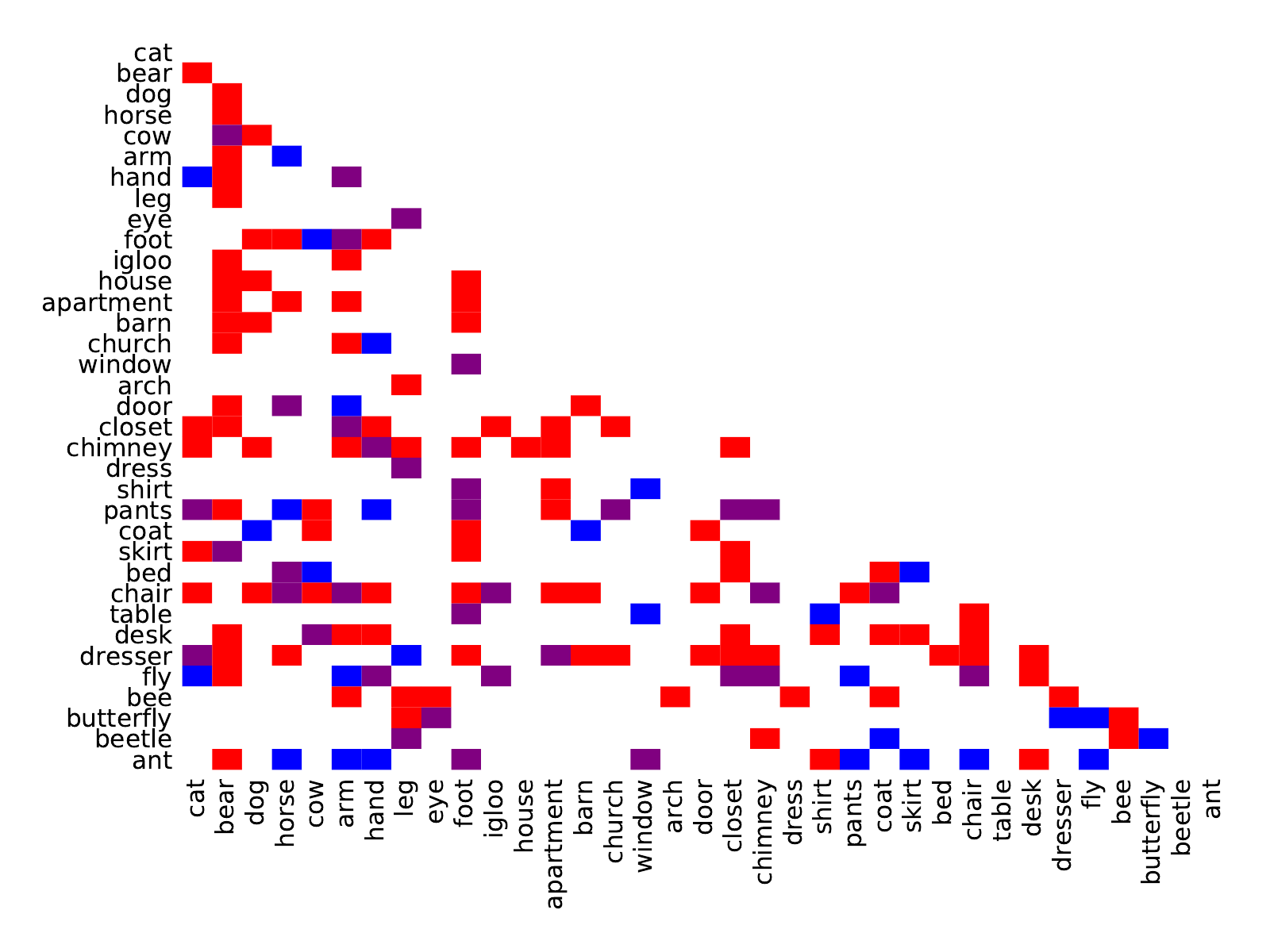}
        \caption{Mismatched word pairs for subject 1: 25 features model (red) vs experiential model (blue). In purple, word pairs confused by both models.}
        \label{fig:exp-25F}
\end{figure}

\begin{figure}
        \includegraphics[width=.5\textwidth,trim={0 0 0 1.3cm}]{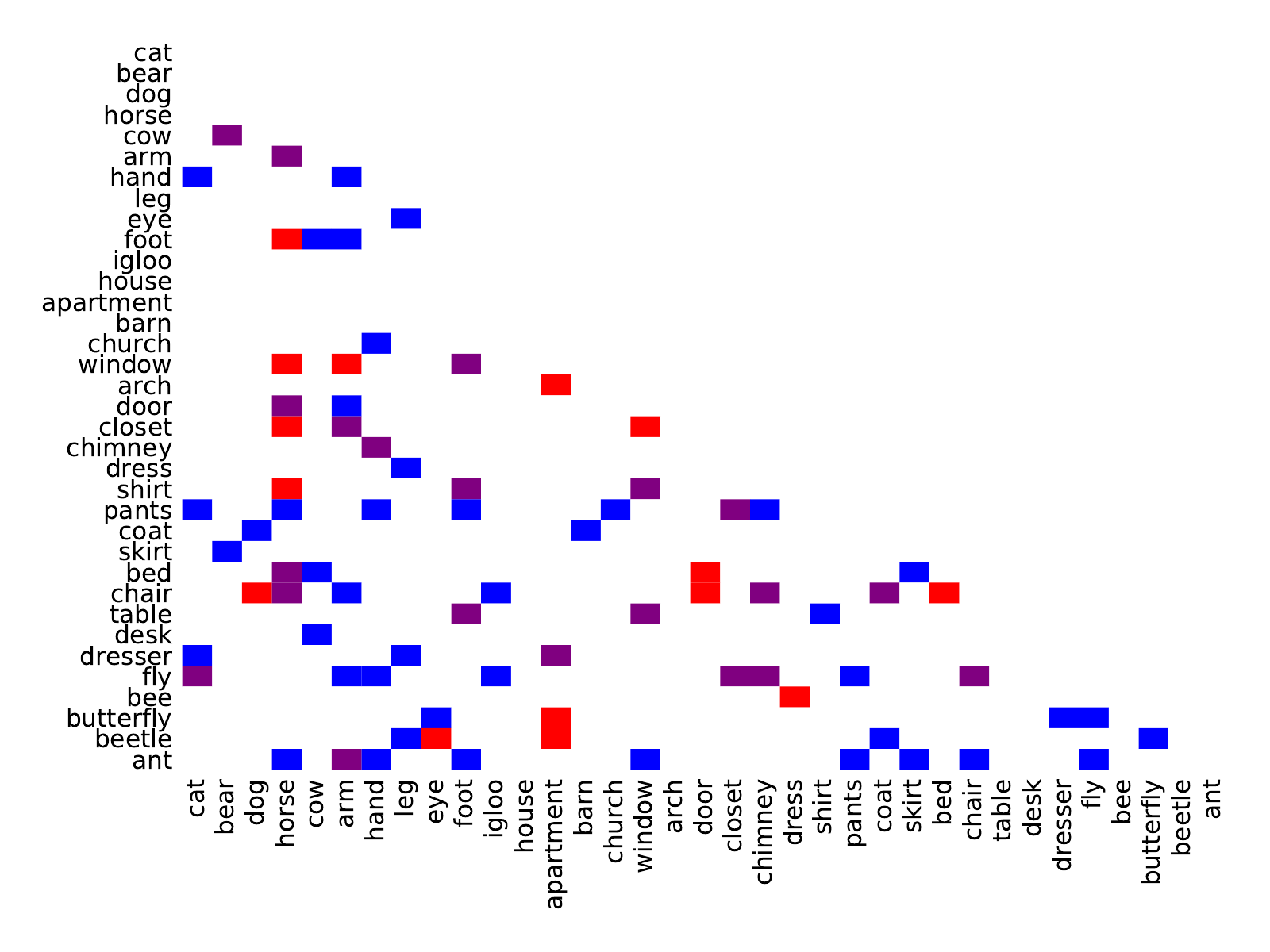}
        \caption{Mismatched pairs for subject 1: dependency based word2vec (red) vs experiential model (blue)}
        \label{fig:exp-dep}
\end{figure}

\begin{figure}
        \includegraphics[width=.5\textwidth,trim={0 0 0 1.3cm}]{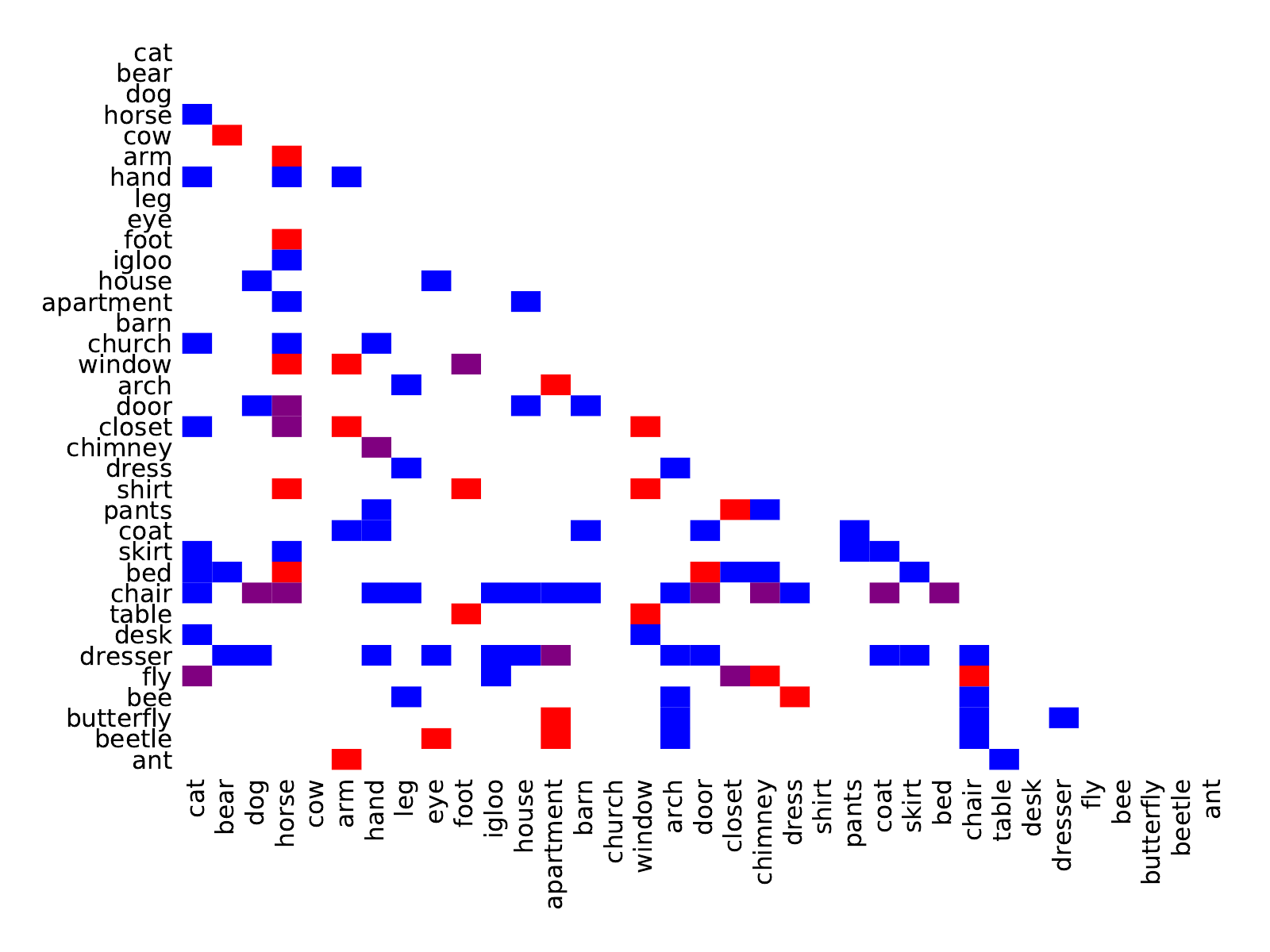}
        \caption{Mismatched pairs for subject 1: dependency based word2vec (red) vs word2vec (blue)}
        \label{fig:dep-word2vec}
\end{figure}

\begin{comment}
\begin{figure*}[t]
    \begin{subfigure}[b]{0.31\textwidth}

    \end{subfigure}
    \hfill
    \begin{subfigure}[b]{0.31\textwidth}

    \end{subfigure}
    \hfill
    \begin{subfigure}[b]{0.31\textwidth}

    \end{subfigure}

    \caption{Difference between mismatched pairs for subject 1}
    \label{fig:mismatched_sub1}
\end{figure*}
\end{comment}

\begin{figure*}[t]
    \begin{subfigure}[b]{0.31\textwidth}
        \includegraphics[width=\textwidth,trim={0 0 0 1.3cm}]{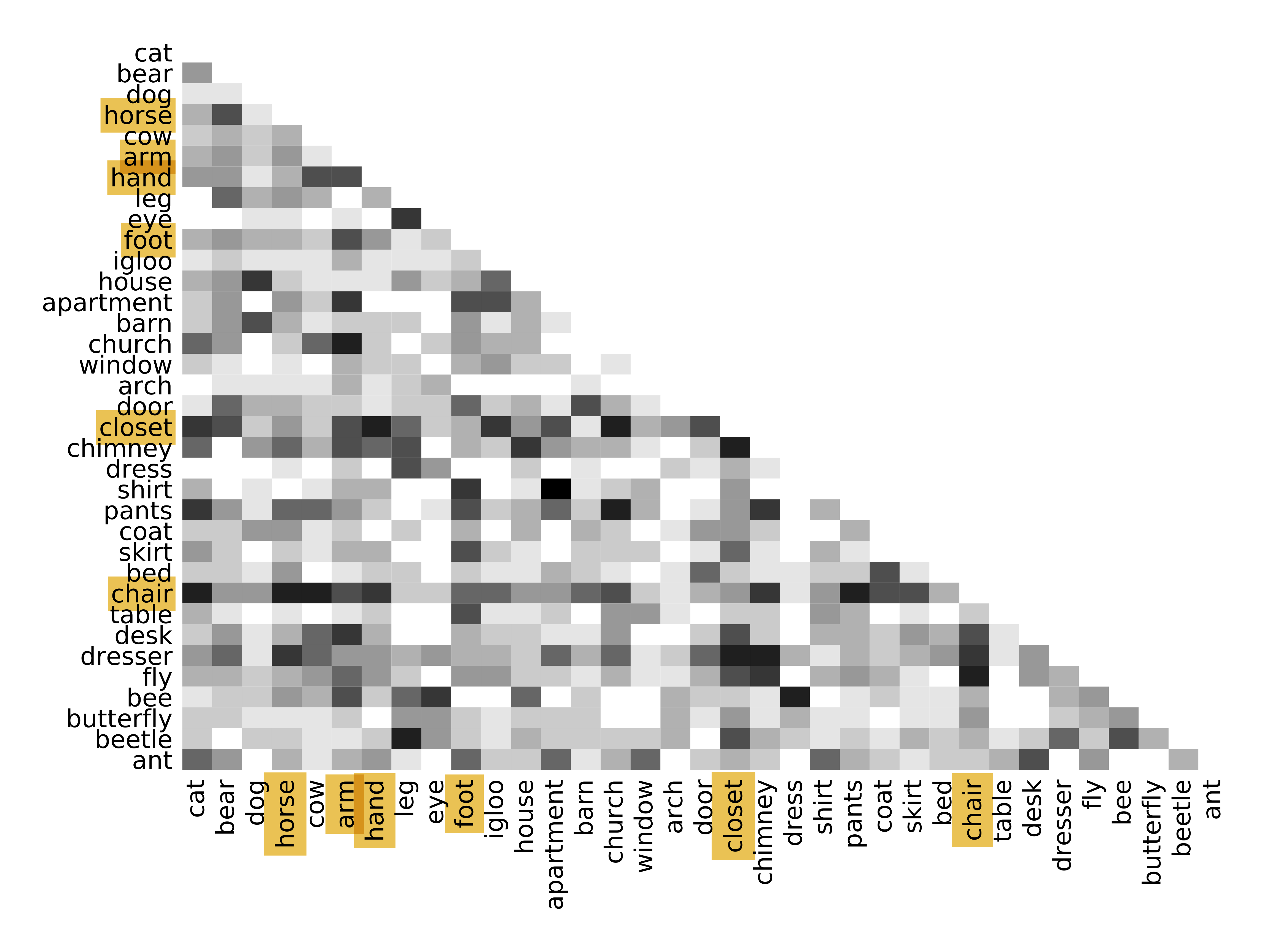}
        \caption{Mismatched pairs for the 25 features model}
        \label{fig:miss25}
    \end{subfigure}
    \hfill
    \begin{subfigure}[b]{0.31\textwidth}
        \includegraphics[width=\textwidth,trim={0 0 0 1.3cm}]{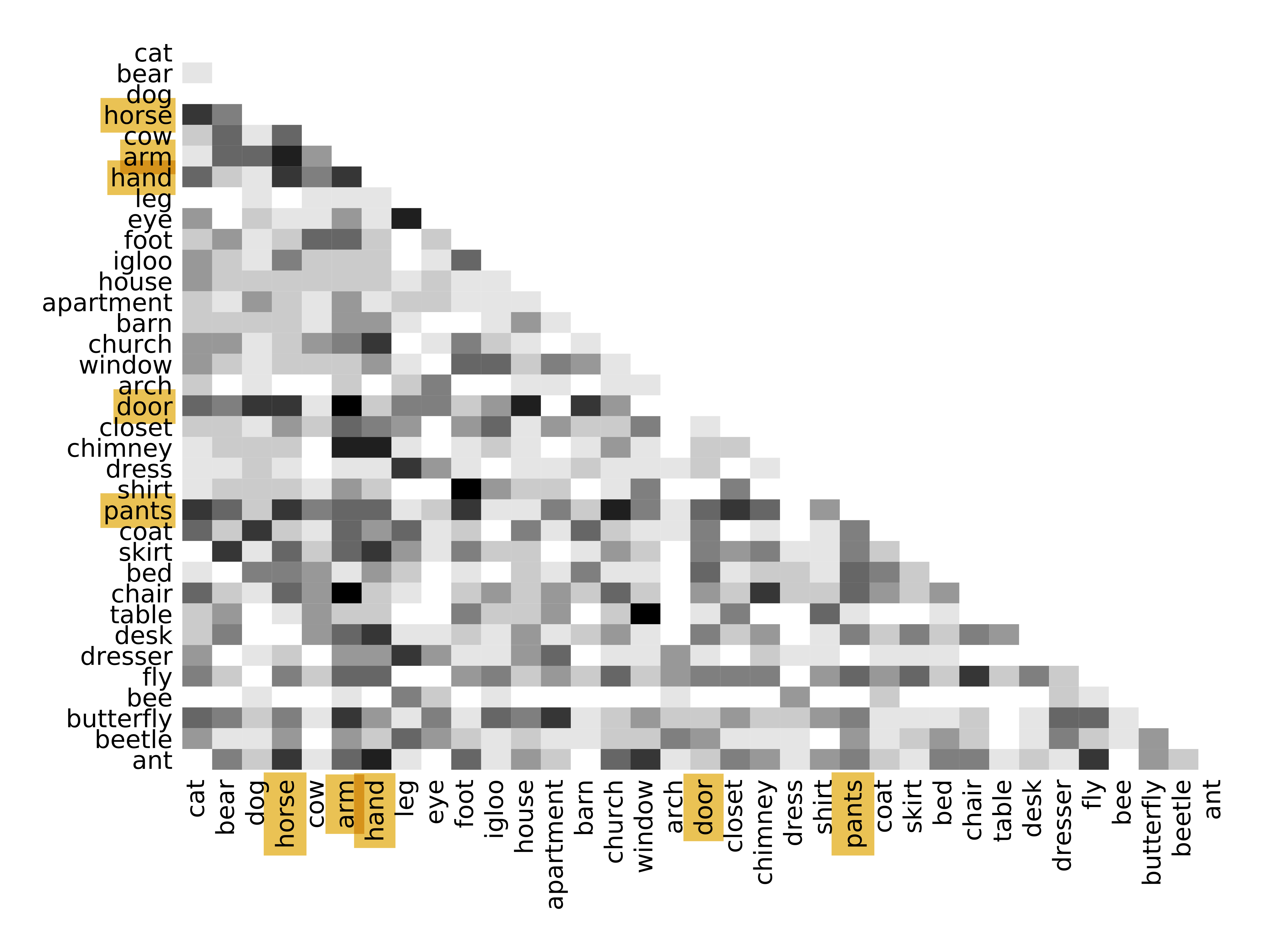}
        \caption{Mismatched pairs for the experiential model}
        \label{fig:missexp}
    \end{subfigure}
    \hfill
    \begin{subfigure}[b]{0.31\textwidth}
        \includegraphics[width=\textwidth,trim={0 0 0 1.3cm}]{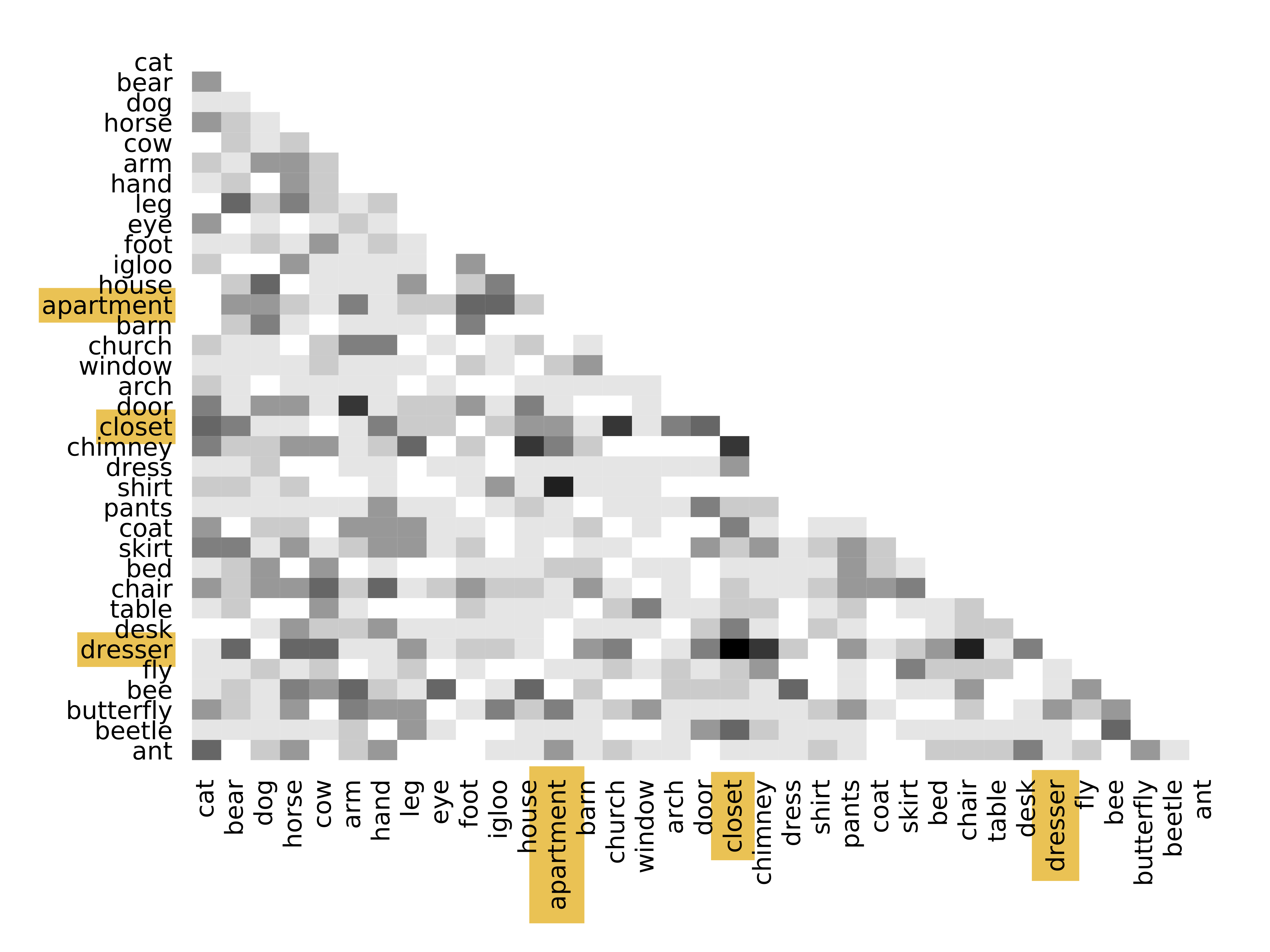}
        \caption{Difference between mismatched pairs\\~}
        \label{fig:miss_25-exp}
    \end{subfigure}
    \caption{Comparing mismatched pairs for the 25 features model and the experiential model averaged over all subjects. Axes are the same as in figure~\ref{fig:exp-25F}.}
    \label{fig:mismatched_25-exp}
\end{figure*}

\begin{figure*}[t]
    \begin{subfigure}[b]{0.31\textwidth}
        \includegraphics[width=\textwidth,trim={0 0 0 1.3cm}]{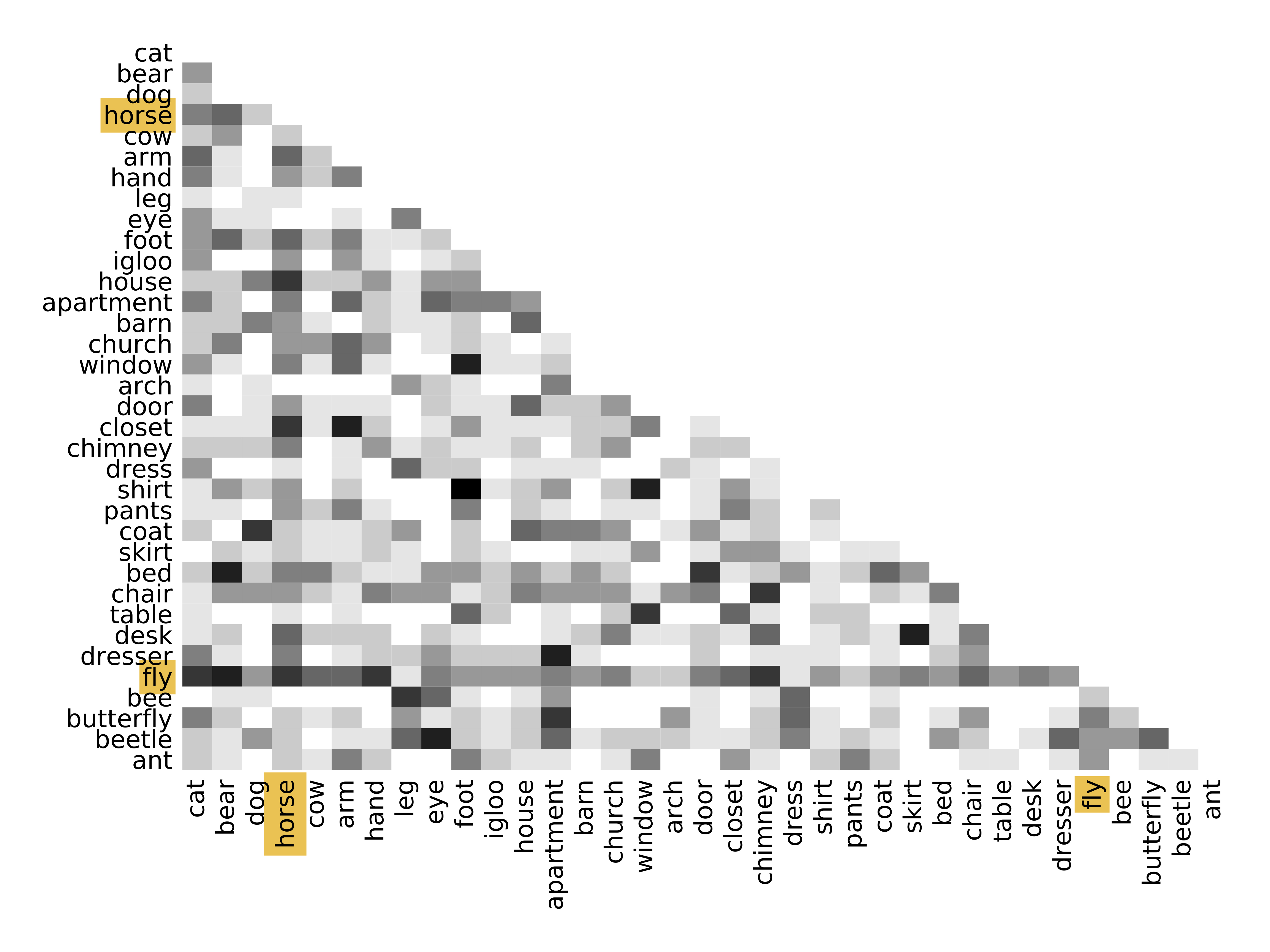}
        \caption{Mismatched pairs for dependency based word2vec}
        \label{fig:missdep}
    \end{subfigure}
    \hfill
    \begin{subfigure}[b]{0.31\textwidth}
        \includegraphics[width=\textwidth,trim={0 0 0 1.3cm}]{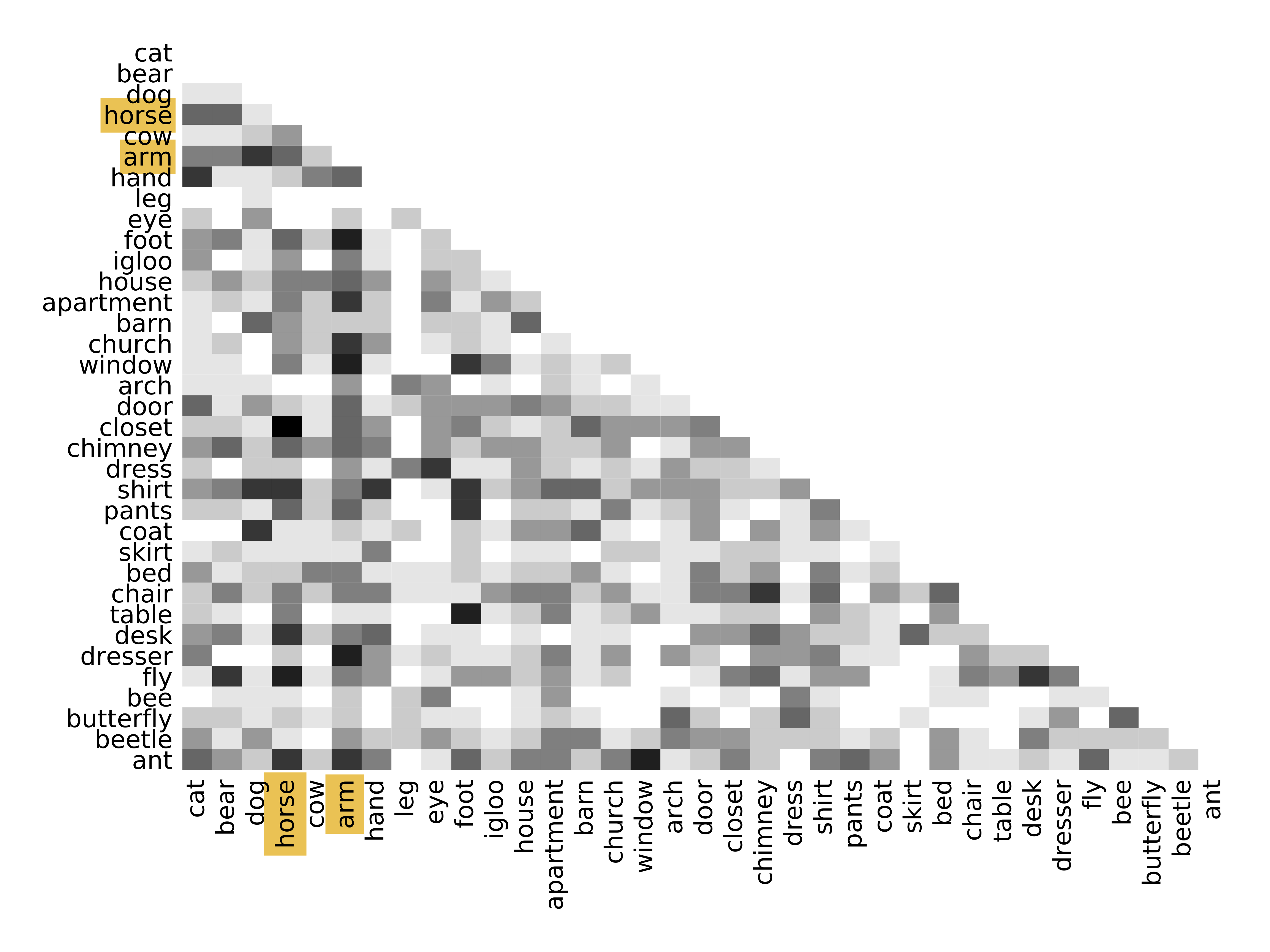}
        \caption{Mismatched pairs for GloVe\\~}
        \label{fig:missglove}
    \end{subfigure}
    \hfill
    \begin{subfigure}[b]{0.31\textwidth}
        \includegraphics[width=\textwidth,trim={0 0 0 1.3cm}]{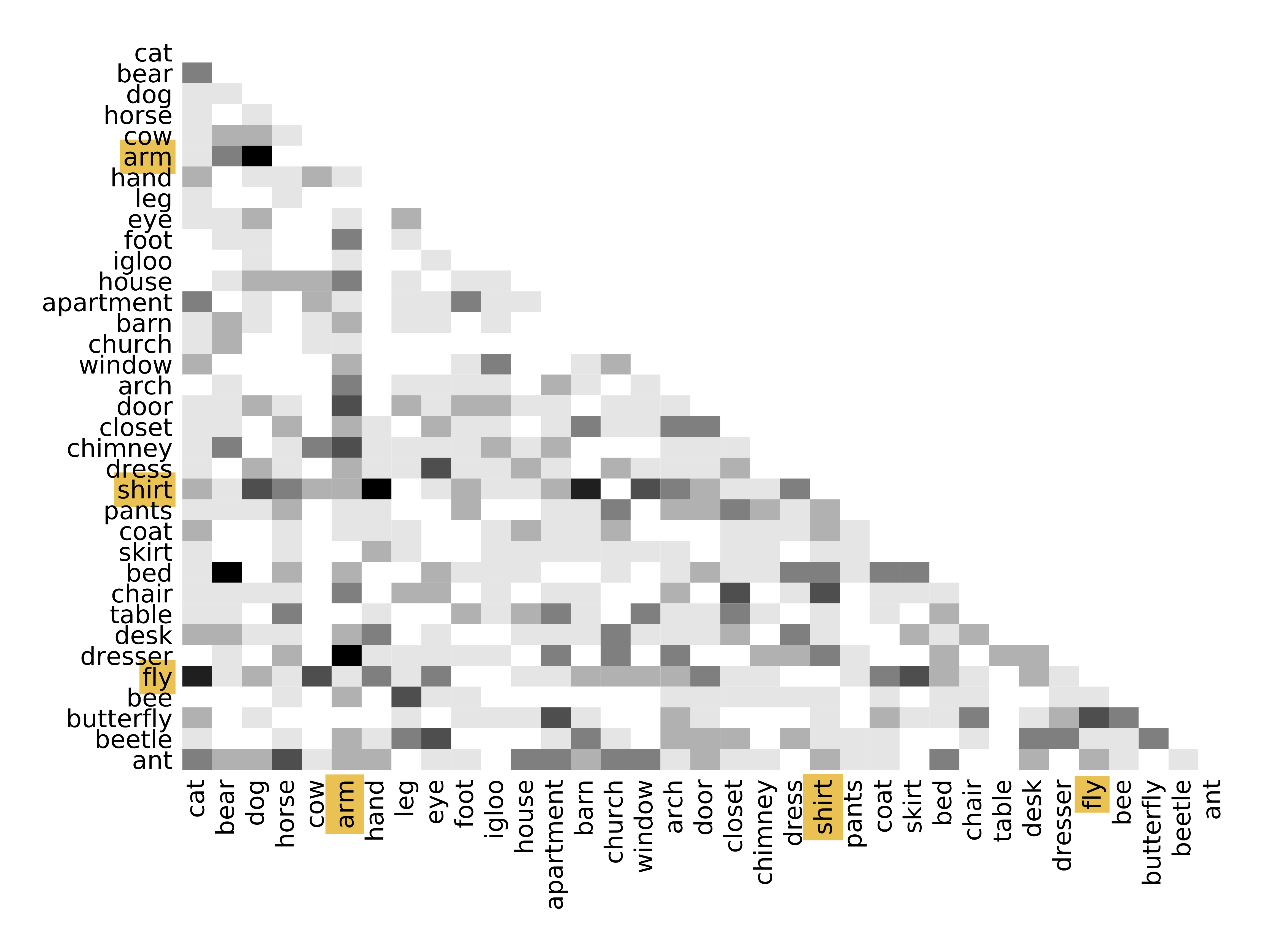}
        \caption{Difference between mismatched pairs\\~}
        \label{fig:missdep-glove}
    \end{subfigure}
    \caption{Comparing mismatched pairs for dependency based word2vec and  GloVe averaged over all subjects}
    \label{fig:mismatched_glove-deps}
\end{figure*}

\begin{figure}[t]
        \includegraphics[width=0.31\textwidth,trim={0 0 0 1.3cm}]{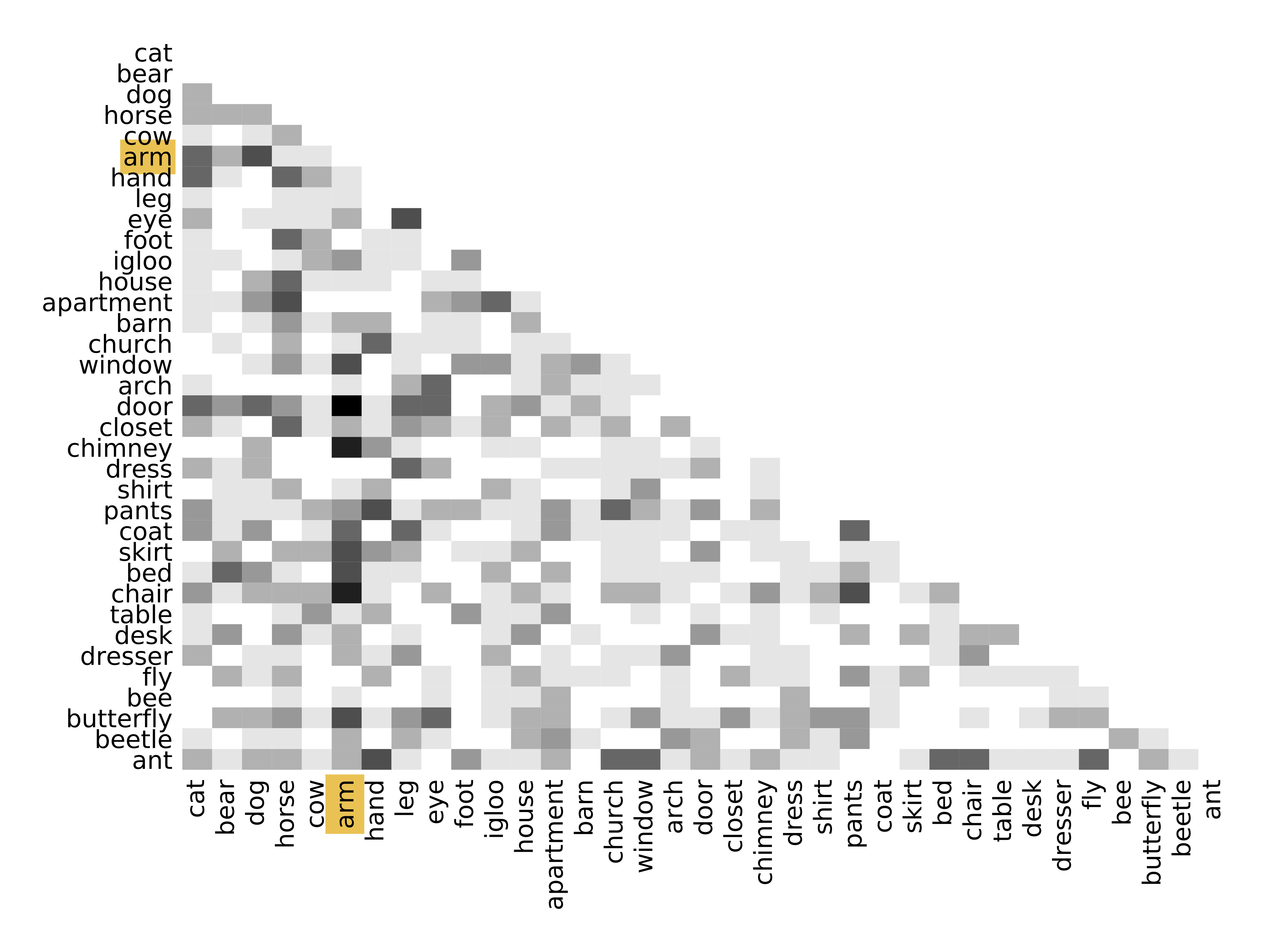}

    \caption{Difference of mismatched pairs for dependency based word2vec and experiential model}
    \label{fig:mismatched_deps-exp}
\end{figure}
\section{\RQThree}
In order to get more insights about the differences between the models, we look into the errors they make.
It is informative to see whether each of these models is good at predicting neural activation pattern for a different group of noun pairs.
We want to test the hypothesis of whether human brain uses different mechanisms for understanding meanings of different categories of words \cite{riddoch1988semantic,caramazza1990multiple,warrington1984category,caramazza1998domain}. 
To investigate this we look into the miss matched noun pairs for each of the word representation models. 
We want to see which are the most confusing noun pairs for each model and measure the overlap between the errors the models make. 
This will reveal if these models are actually encoding different kinds of information.

Figures \ref{fig:exp-25F}, \ref{fig:exp-dep} and \ref{fig:dep-word2vec} show the overlap between mismatched pairs for different models for subject 1. In these plots, the red color corresponds to the first model mentioned in the caption, the blue colour corresponds to the second model and the purple colour indicates the overlaps. 
While there is some overlap between the mistakes of the 25 features model and the experiential model, considerable number of mismatched pairs are not in common between them.
One interesting fact about the 25 features model is that for some specific nouns ie. ``bear'', ``foot'', ``chair'', and ``dresser'', no matter what is its pair, discrimination performance is poor. eg. ``bear'' is not only confused with other animals, but also with some body parts, places and etc. We do not notice similar phenomena for the experiential model. This could be a side effect of using co-occurrence statistics from corpora to learn word representations and could show that for some reason the representations learned for these nouns are not distinguishable from other nouns.
Looking into the noun pair mismatches of the experiential model and the dependency based word2vec in Figure \ref{fig:exp-dep}, again we see a considerable amount of overlap. They both perform equally for discriminating among animals. But the experiential model makes more mistake about ``body parts'' and ``insects''.
Comparing the dependency based word2vec with simple word2vec, in Figure \ref{fig:dep-word2vec} we observe similar patterns to Figure \ref{fig:exp-25F}. As illustrated in the plot, discriminating some words eg. ``chair'' is difficult for  word2vec while it's not the case for dependency based word2vec.  
It seems like both experiential attributes of nouns and the dependency information is helping in learning more distinguishable representations for nouns.

\subsection{25 features vs experiential }
As shown in Figure\ref{fig:Exp1_allinall}, the experiential model performs better than the 25 features model in average. Considering the fact that these two models are reflecting the same underlying theory, we might expect that if one of them is more accurate, it can replace the other. However, by looking into the difference between their mismatched pair, Figure \ref{fig:mismatched_25-exp}, we observe that the mistakes these two models make are not completely overlapping: the nouns `arm' and `hand' are difficult to discriminate for both models, while `chair' and `house' are among the nouns with most mistakes for the 25 features model, and `horse' and `door' for the experiential model. For both models, most mismatches are in the category of body parts.

\subsection{GloVe vs Dependency-based word2vec}
We also compare the mismatch pairs for GloVe and dependency based word2vec as the two neural models that achieve the highest accuracies in Figure \ref{fig:mismatched_glove-deps}. 
These two models are different both in the richness of the information they use to learn word representations, and also the way they use this information. In glove, the model is trained based on the global co-occurance of words whereas in word2vec word representations are learned based on the context of the words for each example locally. 
For GloVe, similar to the 25 features model and the experiential model, `arm' is one of the hardest to discriminate nouns. But the `body parts' category is not as confusing as for the experience based models. For the dependency-based word2vec, the patterns of errors are somehow different and the most difficult word seems to be `fly'. This is because `fly' can be either verb and noun, and since it is more frequent as a verb, the dependency-based model is learning the representation of its verb form. For GloVe, this is not very problematic because it is only based on co-occurrence counts, thus an average representation is learned.
In general, despite the fact that these two models are based on different assumptions their mismatches have more overlap than for the two experiential models. This may be a side effect of the fact that they both make fewer mistakes.

\subsection{Experiential vs Dependency-based word2vec}
The mismatched pairs of the experiential model and the dependency based word2vec and their difference is illustrated in Figure \ref{fig:mismatched_deps-exp}. The experiential model seems to have less prediction accuracy for noun pairs in the same category.

\section{\RQTwo}

\begin{figure}
    \centering
    \includegraphics[width=0.5\textwidth]{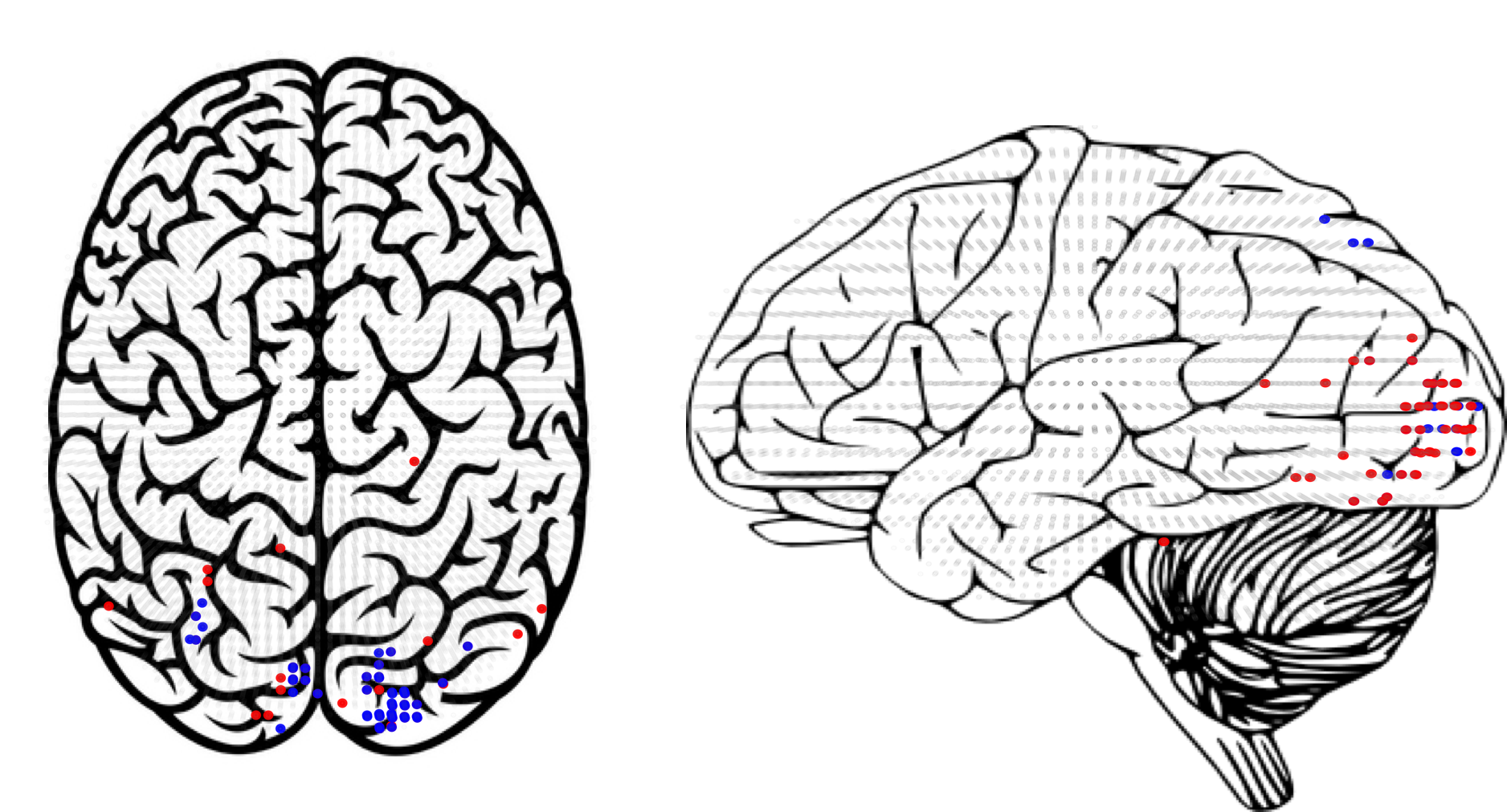}
    \caption{Most predictable voxels for dependecy based word2vec(red) and the experiential model(blue)}
    \label{fig:voxel_1}
\end{figure}

\begin{figure}
    \centering
    \includegraphics[width=0.5\textwidth]{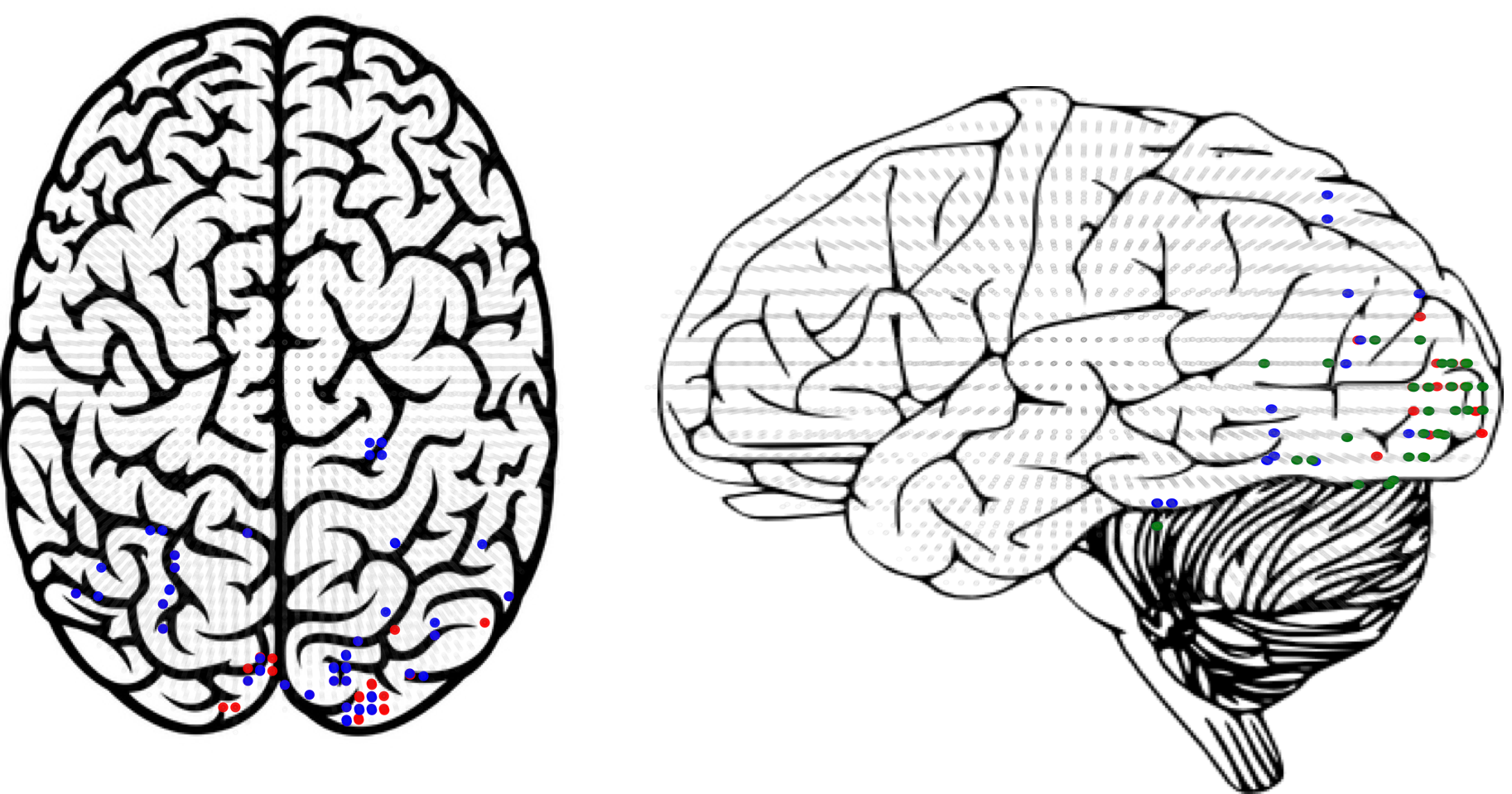}
    \caption{Most predictable voxels for dependecy word2vec(red) and word2vec(blue). Green dots are among the top 50 voxels of both models.}
    \label{fig:voxel_2}
\end{figure}

Each of the computational models of word representation we have employed to predict brain data is based on modelling different aspects of words meanings. Now we want to investigate if our brain is doing a combination of all these mechanisms and different groups of voxels in the brain are responsible for processing each aspect? One way to test this is to look into the predictability of different voxels with each of these models. For this purpose, we have identified the top 50 most predictable voxels for each model. In Figure \ref{fig:voxel_1} you can see the 50 most predictable voxels for dependency-based word2vec and the experiential model.
In Figure \ref{fig:voxel_2} you can see the 50 most predictable voxels for dependency-based word2vec and simple word2vec. The green colour indicates the common top voxels between the two models.
From these figures, we can see that there is a lot more overlap between the dependency based word2vec and word2vec, compared to the experiential model.

\paragraph{A Mixed Model}

If each model is good at predicting the neural activation pattern for a different group of nouns/different groups of voxels, theoretically, it is possible to build a better model using an integrated model. In other words, we should be able to improve the accuracy of predicting neural activation patterns by employing a combined model. 
We conduct a new experiment by integrating the dependency based word2vec as a neural corpus-based word representation with the
experience based models, ie the 25 verbs model and the experiential model. We expect the performance of the model to be a little bit higher than the dependency based word2vec. Our results indicate that combining the dependency based word2vec with the experiential model linearly doesn't lead to an improvement in the accuracy over the limited set of words available in the experiential model. However, linearly combining the 25 feature model with the dependency based word2vec leads to an accuracy of $82$ percent over the 60 nouns, which is $2$ percent higher than the accuracy of the dependency-based model.

\section{Discussion and Conclusion}

Based on our systematic comparison, we can conclude that the deep learning models for learning word representations fit very well with brain imaging data. The existing models, like dependency based word2vec, are already beating the experiential word representation models that are particularly designed for the brain activation decoding tasks. Moreover, comparing the results of learning the mappings from words to brain activations and vice versa, convinces us that it is important to study the performance of the models in both directions to really understand what kind of information is encoded in the neural activation patterns for words. 

Looking into the details of the performance of these models, it turned out that each of them makes different kinds of mistakes.
One of the main problems of the corpus based distributional models that we have applied is that they do not account for different senses of the words. Hence, the representations they learn for words with more than one sense can be noisy and biased toward the most frequent sense. 
Taking the differences between the models into account, we build a model that combines the experience based word representation model with the dependency based word2vec. 
By linearly combining the 25 features model with the dependency-based model we are able to achieve a higher accuracy on the brain activation prediction task. We think it is possible to build new models upon the dependency based word2vec which also encode experiential information. One possible approach to achieve this goal is to train word embedding models in a multi-task learning framework with the downstream tasks that reflect different types of real-life experiences in addition to language modelling tasks.

In addition, in order to have a better understanding of the differences between different word representation models, we need to do a further analysis to answer the question \RQTwo

\section{Acknowledgement}
The work presented here was funded by the Netherlands Organisation for Scientific Research (NWO), through a Gravitation Grant
024.001.006 to the Language in Interaction Consortium.

\bibliographystyle{acl_natbib}
\bibliography{refs} 

\end{document}